\documentclass[sigconf]{acmart}

\usepackage{booktabs} % For formal tablesk
\usepackage[ruled,vlined,linesnumbered]{algorithm2e}
\usepackage{array}
\usepackage{multirow}

\begin{document}

\copyrightyear{2017} 
\acmYear{2017} 
\setcopyright{acmlicensed}
\acmConference{SAICSIT '17}{September 26--28, 2017}{Thaba Nchu, South Africa}\acmPrice{15.00}\acmDOI{10.1145/3129416.3129420}
\acmISBN{978-1-4503-5250-5/17/09}

\title{Automated Classification of Text Sentiment}

\author{Emmanuel Dufourq}
\affiliation{%
  \institution{African Institute for Mathematical Sciences}
  \streetaddress{Maths \& Applied Maths, University of Cape Town}
  \city{Cape Town, South Africa} 
}

\email{edufourq@gmail.com}

\author{Bruce A. Bassett}
\affiliation{%
  \institution{African Institute for Mathematical Sciences}
  \streetaddress{South African Astronomical Observatory}
  \city{Maths \& Applied Maths, University of Cape Town} 
  \state{Cape Town, South Africa} 
}

\email{bruce.a.bassett@gmail.com}

\begin{abstract}
The ability to identify sentiment in text, referred to as sentiment analysis, is one which is natural to adult humans. This task is, however, not one which a computer can perform by default. Identifying sentiments in an automated, algorithmic manner will be a useful capability for business and research in their search to understand what consumers think about their products or services and to understand human sociology. Here we propose two new Genetic Algorithms (GAs) for the task of automated text sentiment analysis. The GAs learn whether words occurring in a text corpus are either sentiment or amplifier words, and their corresponding magnitude. Sentiment words, such as 'horrible', add linearly to the final sentiment. Amplifier words in contrast, which are typically adjectives/adverbs like 'very', multiply the sentiment of the following word. This increases, decreases or negates the sentiment of the following word. The sentiment of the full text is then the sum of these terms. This approach grows both a sentiment and amplifier dictionary which can be reused for other purposes and fed into other machine learning algorithms. We report the results of multiple experiments conducted on large Amazon data sets. The results reveal that our proposed approach was able to outperform several public and/or commercial sentiment analysis algorithms.
\end{abstract}




\maketitle

\section{Introduction}

The amount of data collected from users around the world and stored for posterity has skyrocketed over the past decade as websites such as Twitter, Amazon and Facebook have facilitated the publication and aggregation of micro opinion pieces that allow individuals to record their sentiments towards things, people and events. This data is clearly of value to researchers, organizations and companies to understand sentiment both as individuals and on average, and as well as to identify trends. The automated detection of emotions and attitudes towards a particular subject, event or entity is what we will call sentiment analysis \cite{Pang:2008:Opinion, liu:2015:sentiment}. Sentiment analysis has been applied to many problem domains; for instance, determining sentiments of consumers towards products, or mining social media to gain an understanding of the public's opinion on matters such as corruption \cite{liu:2015:sentiment, Pang:2008:Opinion, Ravi:2015:ASurvey}.

For adult humans, interpreting the underlying emotions in text is usually performed unconsciously and with apparent ease. We are able to recognize emotions in emails, sentiments in our social media feed and appreciate the subtle nuances of conflicting views in novels. Nevertheless, even for humans text can be notoriously easy to misinterpret. For machines, on the other hand, sentiment analysis is highly non-trivial. From the year 2000 onwards, a number of researchers have begun contributing towards the field of sentiment analysis \cite{Pang:2008:Opinion}. This area of research is highly active, increasingly so, due to the vast amount of digital information available, and the amount of sentiments expressed online. With the rapid increase in computational power available in the recent years and the extreme amount of data available online, it is clear that developing novel sentiment analysis methods will be beneficial to organizations in order to enable them to understand what the public feels about their products and services.

In this study, two genetic algorithms (GA) were proposed for text sentiment analysis. The proposed approach optimizes the sentiments of the words in order to correctly classify as much data as possible. This research proposes a new way of representing words, as either a sentiment or an amplifier word, whereby amplifier words intensify the sentiments in sentences. The words are combined to form mathematical expressions in order to determine whether or not a given sentence is positive or negative. The following section describes the GA which was used to create the models which perform sentiment analysis.

%---------------------------
% Genetic Algorithm
%---------------------------
\section{Genetic Algorithm}\label{sec: gas}

A GA \cite{Goldberg:1989:GeneticAlgorithms} is an evolutionary algorithm \cite{Eiben:2003:IntroTo} inspired by  ``survival of the fittest'' in nature that can be used to solve optimization problems. A GA evolves a population of chromosomes which are made up of several genes. The size of the population is a user-defined parameter. Each chromosome represents a candidate solution to the optimization problem. Each chromosome is evaluated in order to determine how successful it is at solving the optimization problem. The evaluation is obtained by computing the \textit{fitness} of each chromosome. For a maximization problem, a chromosome with a higher fitness is considered as a better one, whereas a chromosome with a smaller fitness is considered as a weaker one. 

This study implements GAs to optimize the correct classification (positive or negative) of short pieces of text given words of unknown sentiment in the text. While some sentiment algorithms make use of large dictionaries of words with associated sentiment values \cite{Musto:2014:AComparison}, the GAs we propose can learn the type and associated sentiment values of words; although the GA can make use of sentiment dictionaries if desired. This is appropriate if there is little training data. 

Algorithm \ref{algo:ga-algo} illustrates the pseudocode for a GA. An initial population of chromosomes is randomly created in step 2, and each chromosome is evaluated in step 3 to determine if a solution to the optimization problem exists from the initial population. In step 5, the algorithm enters into a generational loop until the maximum number of generations is met, or until a solution to the optimization problem is found. The maximum number of generations is a user-defined parameter. 	

\begin{algorithm}

\SetKwData{Gen}{generation}
\SetKwData{Genmax}{generation\_max}
\SetKwInOut{Input}{input}

\Input{\Genmax : maximum number of GA generations}

	\Begin{
	
	Create an initial population of chromosomes.
	
	Evaluate the initial population.
	
	\Gen $\leftarrow 0$.
	
	\While{\Gen $\leq$ \Genmax}{
	
		\Gen $\leftarrow \Gen+1.$
	
		Select the parents.
		
		Perform the genetic operators.
		
		Replace the current population with the new offspring created in step 8.

		Evaluate the current population.
	
	}

\Return{The best chromosome.}

}
\caption{Genetic algorithm} 
\label{algo:ga-algo} 
\end{algorithm}

%---------------------------
% Related Work
%---------------------------
\section{Related Work}\label{sec: relatedwork}

GAs have been used before in sentiment analysis studies, though not primarily for actual sentiment determination but rather for feature selection and reduction, e.g. \cite{Abbasi:2008:SentimentAnalysis}. There the chromosomes had length equal to the total number of features, and the genes were encoded with a \textit{0} or a \textit{1} depending on whether or not that particular feature was to be used or not. The GA optimized which features to use from the original set, and a SVM classifier was then applied to that feature set in order to train and predict the reviews. Genes were encoded in a similar manner for feature selection with the ultimate goal of reducing the number of features in the study of Kalaivani and Shunmuganathan \cite{Kalaivani:2015:FeatureReduction}.

GAs were also used to optimize features in several other studies, such as that of Paramesha and Ravishankar \cite{Paramesha:2013:Optimization} which used a GA in order to allocate weights to features. Govindarajan \cite{Govindarajan:2013:SentimentAnalysis} proposed an ensemble approach using Naive Bayes and a GA. Smith \cite{Smith:2010:UsingGenetic} proposed the use of GA to reduce the number of features as did Acampora and Cosma \cite{Acampora:2014:AHybrid}.

Carvalho et al. \cite{Carvalho:2014:AStatistical} present a novel GA approach whereby a fixed chromosome is split in two parts, a positive and a negative part. A set of 25 positive and 25 negative words were seeded into the algorithm. Their approach attempts to find which of those words should be added into the respective parts of the GA chromosomes in order to maximize the accuracy of classifying Twitter tweets. A chromosome is then evaluated using a distance measure based on the words in the tweets in relation to the words in the chromosome. Thus, for example, if a particular chromosome is evaluated on some tweet, and the words in the tweet are considered to be nearer (based on the distance measure) to the positive words in the chromosome than to the negative, then the tweet is classified as positive. 

Das and Bandyopadhyay \cite{Das:2010:Subjectivity} make use of a GA for subjectivity detection. Even though this area of research does not deal with sentiments, the research is aligned. Ten features were chosen and a number of predetermined values were assigned to each feature. An example of two features used were parts-of-speech and SentiWordNet values. The former takes up to 45 possible parts-of-speech values; and latter 2 values, positive or negative. The aim behind the research was to optimize the best set of features.

By contrast, the rationale behind the present study is not to propose a new GA feature selection method; instead, the focus is to propose a GA that determines the sentiment of reviews without making use of a feature set. Furthermore, our approach treats each individual piece of text with a sentiment as a mathematical formula made up of unknown variables corresponding to each word in the text. Thus, the goal is to use a GA to simultaneously solve for the unknown variables as a step towards correctly predicting the total sentiment of a piece of text.

%---------------------------
%CLASSIFICATION VALUE PAIR
%---------------------------
\section{Classification-Value Pair}\label{Classification-value}

In our study, each word is assigned both a `classification' and a `value' that we call a classification-value pair in the form `\textit{classification:value}'. Classifications take on one of two types, namely either \textit{sentiment} or \textit{amplifier}. Intuitively this captures the difference between words that carry sentiment directly (e.g. `horrible', `sad', `wonderful') and adjectives/adverbs that modify the sentiment of the following word (e.g. `very', `not', `little'). In addition to this classification every word is given exactly one value associated with that classification, taken from this list:

\begin{itemize}
\item Sentiment $\in$ \{\textit{-1.0}, \textit{0.0}, \textit{1.0}\}
\item Amplifier $\in$ \{\textit{0.5}, \textit{1.0}, \textit{1.5}\}
\end{itemize}

For this classification-value pair, a sentiment value of -1, 0 and 1 represents a negative, neutral and positive sentiment respectively. The three values for the amplifier represent different intensification values, i.e. a value of 1.5 is a larger amplification than a value of 0.5. These values were selected by conducting various preliminary runs.

A word is referred to as an unknown word if its classification-value is not known. Examples of three classification-value pairs are: \textit{sentiment:1.0}, \textit{amplifier:0.5}, and \textit{sentiment:-1.0}.

The goal of this study is to optimize and determine the classification-value pairs for certain unknown words within a data set, given that, a number of words already have known classification-value pairs. The words which already have a known classification-value pair are stored in a dictionary. Words in a dictionary do not have to be optimized and their classification-value pairs are never altered. 

In this study we use two dictionaries, one for sentiment words and one for amplifier words. Known sentiment words were added into the sentiment dictionary, and similarly, known amplifier words were added into the amplifier dictionary. This was done to provide seeds as to guide the GA to converge to the correct solutions. Furthermore, the proposed algorithm can extend these dictionaries in order to create a sentiment lexicon. Details regarding which dictionaries were used are provided in section \ref{results}.

%---------------------------
%PROPOSED GA
%---------------------------
\section{Proposed Methods for Optimizing Classification-Value Pairs}\label{Proposed}

This section describes the use of machine learning in order to optimize the classification-value pairs for the unknown words in the sentences of a data set. We propose Genetic Algorithm for Sentiment Analysis (\textit{GASA}). Each aspect of the GA is explained in terms of how it has been adapted for \textit{GASA} in the following subsections.

%---------------------------
%GASA REPRESENTATION
%---------------------------
\subsection{GASA chromosome representation}\label{sub:chromoRep}

Each gene within a chromosome is made up of the classification-value pair for an unknown word (not in the sentiment or amplifier dictionary). The length of the chromosome is equal to the number of unknown words in the training corpus. The classification for each unknown word corresponds to a gene in the chromosome, and thus the classification of unknown words: $word_{1},word_{2}, word_{3}$, \ldots,$word_{n}$ is mapped to: $gene_{1},gene_{2},gene_{3},$\ldots$,gene_{n}$ --- where $n$ represents the number of unknown words in a training corpus. This mapping is never changed.

\begin{figure}[!ht]
  \centering
          \includegraphics[width=0.45\textwidth]{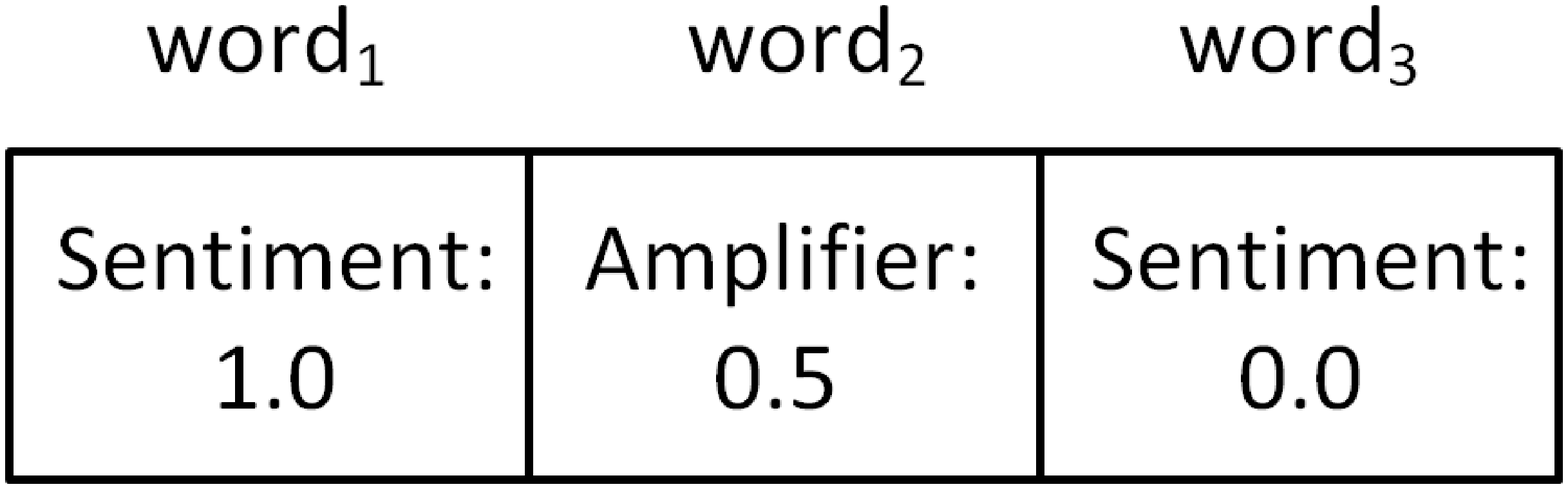}
  \caption{Example of a \textit{GASA} chromosome.}
\label{fig:exampleChromo}
\end{figure}

In order to illustrate the chromosome representation, suppose there are three unknown words: $word_{1},word_{2},word_{3}$. Figure \ref{fig:exampleChromo} illustrates an example of a candidate chromosome of length 3. The illustrated chromosome corresponds to the following classification:

\begin{itemize}
\item $word_1$ in gene position 1 is classified as a \textit{sentiment} word with a value of $1.0$
\item $word_2$ in gene position 2 is classified as an \textit{amplifier} word with a value of $0.5$
\item $word_3$ in gene position 3 is classified as a \textit{sentiment} word with a value of $0.0$
\end{itemize}

%---------------------------
%GASA INIT POP GEN
%---------------------------
\subsection{GASA initial population generation}

Prior to creating the initial population, the unknown words have to be input into the GA. The initial population size is set to the same value as the user-defined population size. Suppose the population size is \textit{n}, then \textit{n} chromosomes are  created during the initial population generation. Each chromosome has a fixed length which is set to the number of unknown training words. The pseudocode for creating a chromosome is presented in algorithm \ref{algo:create-chromosome}. The genes which make up the chromosome are created by randomly selecting either a sentiment or an amplifier classification and assigned a value randomly as described in section \ref{Classification-value}.

\begin{algorithm}

\SetKwData{Size}{size}
\SetKwInOut{Input}{input}

\Input{\Size : the number of unknown words}

	\Begin{
	
	Initialise the length of the chromosome to \Size

	\For{each gene in the chromosome}{

		Randomly select a classification type.

		Randomly select a corresponding value for the classification type previously obtained in step 4.

	}

}
\caption{Creating a chromosome.} 
\label{algo:create-chromosome} 
\end{algorithm}

%---------------------------
%GASA EVALUATION
%---------------------------
\subsection{GASA chromosome evaluation}\label{sub:chromoEval}

A chromosome has to be evaluated in order to determine how good it is at solving the optimization problem (namely how well it predicts the overall sentiment of a piece of text). Every chromosome is evaluated on each instance in the data set. In this study, an instance corresponds to text. 

Assume that chromosome \textit{c} is being evaluated. Chromosome \textit{c} is applied to every instance in the data set, and each word within the instances is examined in order to obtain its classification-value pair. Assume that chromosome \textit{c} is evaluating instance \textit{i}, whereby the text for instance \textit{i} is made up of the following words: $w_{1}, w_{2}, w_{3},$ $\dots,$ $w_{n}$, and \textit{n} denotes the length of instance \textit{i}.

If a word $w_{i}$ from instance \textit{i} is in the sentiment dictionary, then $w_{i}$ is classified as a sentiment word, and its corresponding value is retrieved from the sentiment dictionary. Similarly, if $w_{i}$ is in the amplifier dictionary, then $w_{i}$ is classified as an amplifier word, and its corresponding value is obtained from the amplifier dictionary. If however, $w_{i}$ is unknown, then its classification-value pair is retrieved from chromosome \textit{c}.

Once the classification-value pair for every word in an instance of data has been obtained, these classification-value pairs are converted into a mathematical expression in order to obtain the sentiment for the instance. The mathematical expression is evaluated sequentially from left to right. Algorithm \ref{algo:eval-sentence} presents the pseudocode to evaluate expressions. Amplifier words boost the sentiment words, whereas the sentiment words accumulate each other. If the final word is an amplifier, then that value is simply added onto the result. A positive output denotes a positive sentiment, and a negative output denotes a negative sentiment.

The fitness of a chromosome is determined as the total number of instances for which the sentiment output by the chromosome is equal to the correct sentiment from the data set. Assume that some data set has sentences $s_{1} , s_{2} ,$ and $s_{3}$, and these have correct sentiments of \textit{positive, negative, positive} respectively. If some chromosome evaluates each sentence to: \textit{negative, negative, negative}, then the fitness of that chromosome is one, since, it only correctly classified the second sentence.

\begin{algorithm}

\SetKwData{Sen}{sentence}
\SetKwData{Sentiment}{sentiment\_count}
\SetKwData{Amplifier}{amplifier\_count}
\SetKwData{Word}{word}
\SetKwInOut{Input}{input}
\SetKwInOut{Output}{output}

\Input{\Sen : the sentence to be evaluated}
\Output{The sentiment for the evaluated sentence}

\Begin{

\Sentiment $\leftarrow$ 0

\Amplifier $\leftarrow$ 0

\For{each \Word in the \Sen}{

	\If{\Word is an amplifier}{

		\Amplifier $\leftarrow$ \Amplifier + \Word's amplifier value

	}\Else{

		\If{\Amplifier is non-zero}{

			\Sentiment $\leftarrow$ \Sentiment + \Amplifier $\times$ \Word's sentiment value

		}\Else{

			\Sentiment $\leftarrow$ \Sentiment + \Word's sentiment value

		}

	}

}

\If{\Amplifier is non-zero}{
	\Sentiment $\leftarrow$ \Sentiment + \Amplifier
}

\Return{\Sentiment}

}
\caption{Pseudocode for arithmetically evaluating a sentence.} 
\label{algo:eval-sentence} 
\end{algorithm}

%---------------------------
%PARENT SELECTION
%---------------------------
\subsection{GASA parent selection}\label{sub:parentSelection}

Parent selection methods are used to obtain parents from the current population of chromosomes. These parents are used by the genetic operators in order to create offspring. A single parent is obtained when the parent selection method is executed. Once a chromosome has been chosen to be a parent, the selection method can  select that particular chromosome again. Three common parent selection methods are fitness-proportionate, rank and tournament selection \cite{Blickle:1996:AComparison}. For this study, tournament selection was used given that it was shown to be a successful method by Zhong et al. \cite{Zhong:2005:ComparisonOfPerformance}.

Algorithm \ref{algo:tournament-selection} presents the pseudocode for the tournament selection. This selection method has one user-defined parameter, namely, the tournament size. Let \textit{k} be the tournament size. Tournament selection randomly selects \textit{k} chromosomes from the current GA population, and compares the fitness of each of the \textit{k} chromosomes. The chromosome with the highest fitness is returned as the parent chromosome. If a tie occurs, then a random chromosome is selected to break the tie.

\begin{algorithm}

\SetKwData{Size}{size}
\SetKwData{Current}{current\_best}
\SetKwData{Chromosome}{random\_chromosome}
\SetKwInOut{Input}{input}
\SetKwInOut{Output}{output}

\Input{\Size : size of the tournament}
\Output{The best chromosome which will be used as a parent}

\Begin{

	\Current $\leftarrow$ null
	
	\For{$i\leftarrow 1$ \KwTo \Size}{
		
		\Chromosome $\leftarrow$ randomly select a chromosome from the population

		Evaluate \Chromosome

		\If{fitness of \Chromosome $>$ fitness of \Current}{

			\Current $\leftarrow$ \Chromosome

		}

	}

\Return{\Current}

}
\caption{Pseudocode for tournament selection.} 
\label{algo:tournament-selection} 
\end{algorithm}

%---------------
%GASA GOs
%--------------
\subsection{GASA genetic operators}

Genetic operators are applied to parents in order to exchange genetic material between the parent chromosomes, and to consequently create novel offspring. The two most common genetic operators are mutation and crossover. Their implementation details for this study are described below.

%---------------------------
%GASA MUTATION
%----------------------
\subsubsection{GASA mutation}

\begin{figure}[!ht]
  \centering
          \includegraphics[width=0.40\textwidth]{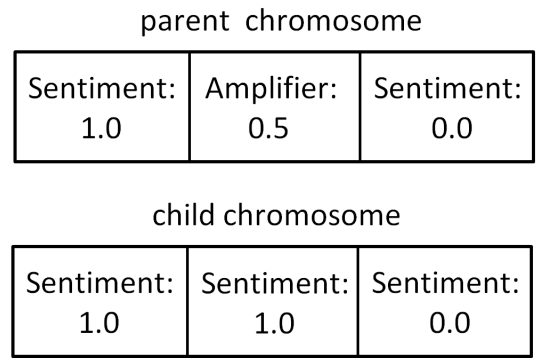}
  \caption{Example of \textit{GASA} mutation. The second gene was selected for mutation and was changed from an amplifier with a value 0.5 to a sentiment with a value of 1.0. The other genes remain unchanged.}
\label{fig:gasamutation}
\end{figure}

The mutation genetic operator makes use of a single parent chromosome. The classification-value for a single gene in the parent is modified to a new one. A user-defined parameter is associated with the mutation operator, namely the mutation application rate. Figure \ref{fig:gasamutation} illustrates the application of the mutation operator on a parent chromosome, and the resulting offspring is illustrated. The second gene in the parent was changed from a classification of ``amplifier'' with a value of 0.5 to a classification of ``sentiment'' with a value of 1.0. 

%---------------------------
%GASA CROSSOVER
%------------------------
\subsubsection{GASA crossover}

The crossover genetic operator exchanges genetic material between two parent chromosomes: $parent_{1}$ and $parent_{2}$, and consequently creates two offspring: $child_{1}$ and $child_{2}$. There are several variations of the crossover genetic operator, such as uniform, one-point and two-point crossover. 

The crossover method we implement randomly selects a position \textit{p} in the range $[0,n]$ --- where \textit{n} denotes the length of the chromosome --- within the parent chromosomes; the same position \textit{p} must be selected within the two parents. Two offspring are created, and all the genes except those at position \textit{p} are copied across to the corresponding offspring without modification. The genes are position \textit{p} are swapped, i.e., the gene in position \textit{p} from $parent_1$ is inserted into position \textit{p} in $child_2$, and similarily, the gene in position \textit{p} from $parent_2$ is inserted into position \textit{p} in $child_1$. 

\begin{figure}[!ht]
  \centering
          \includegraphics[width=0.48\textwidth]{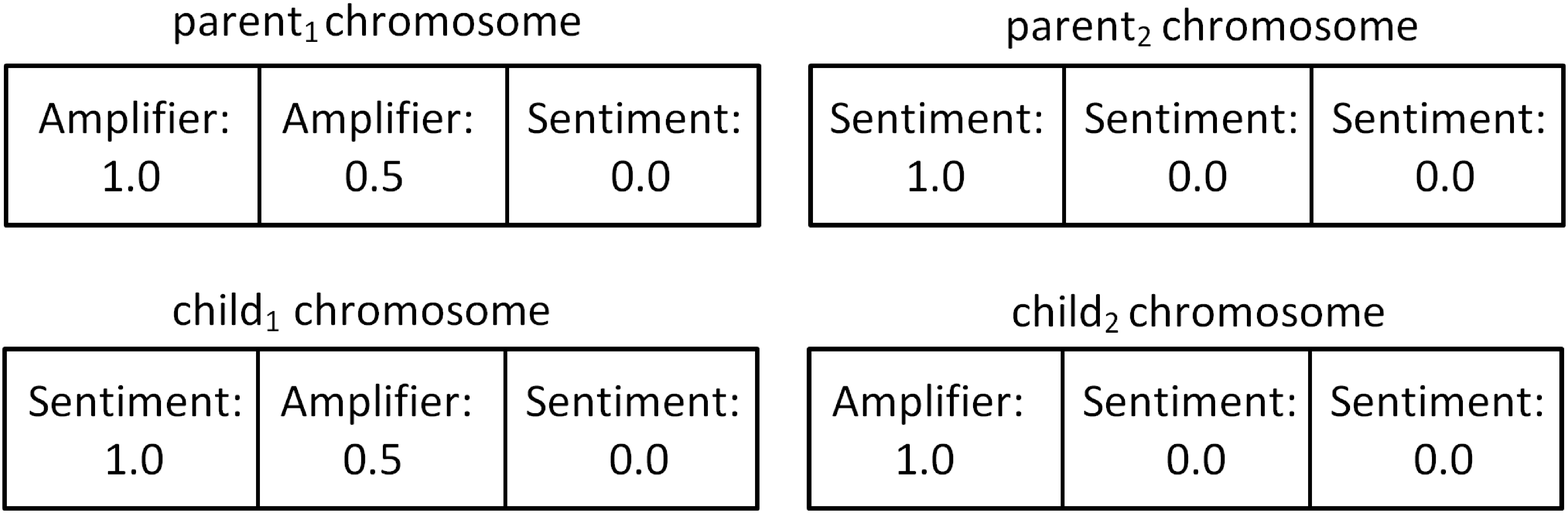}
  \caption{Example of \textit{GASA} crossover. The first gene was swapped between the parents, i.e. the amplifier in the first gene from parent 1 was swapped with the sentiment in the first gene from parent 2. The result of the crossover is observed in the children chromosomes. All of the other genes remain unchanged.}
\label{fig:gasacrossover}
\end{figure}

Figure \ref{fig:gasacrossover} illustrates the application of the proposed crossover operator on two parent chromosomes; the resulting offspring are also illustrated. In this case, the value of \textit{p} was 1, implying that the first gene was swapped amongst the parent chromosomes.

%----------------
% EXPERIMENTAL SETUP
%---------------
\section{Experimental Setup}

This section describes the experimental set up which was used in order to evaluate the performance of \textit{GASA}. \textit{GASA} was programmed in Java and the experiments were conducted at the University of Geneva on the Baobab cluster. 

%----------------
%DATA SETS
%---------------
\subsection{Data sets}

Based on the literature surveyed, there is no consistency in terms of the number of data sets used in previous studies. Furthermore, the total number of reviews also varies from one study to another. For example, Che et al. \cite{Che:2015:SentenceCompression} used a data set containing only 878 reviews, whereas the data set used in the study of Wang et al. \cite{Wang:2010:LatentAspect} had 108891 reviews --- the number of reviews largely differ between these two. Pang et al. \cite{Pang:2002:ThumbsUp} and Govindarajan \cite{Govindarajan:2013:SentimentAnalysis} used 2000 reviews, whereas Acampora and Cosma \cite{Acampora:2014:AHybrid} used 95084. Paramesha and Ravishankar \cite{Paramesha:2013:Optimization} used 2243. Carvalho et al. \cite{Carvalho:2014:AStatistical} used two data sets which had 359 and 1908 reviews. There is no study which guides researchers to a set of recommended benchmark data sets for sentiment analysis. 

For this study, eight data sets were constructed from several Amazon data sets. The Amazon data sets were obtained from Leskovec and Krevl \cite{stanfordSnap} and McAuley et al. \cite{McAuley:2015:Inferring}. Each instance in the Amazon data sets is made up of a short summary and a review which were provided by a user. An example of an instance is as follows:

\begin{itemize}
\item \textbf{summary:} Pittsburgh - Home of the OLDIES

\item \textbf{review:} I have all of the doo wop DVD's and this one is as good or better than the 1st ones. Remember once these performers are gone, we'll never get to see them again. Rhino did an excellent job and if you like or love doo wop and Rock n Roll you'll LOVE this DVD !
\end{itemize}

Four Amazon data sets were randomly selected, namely: Cell Phones and Accessories (Cellphone), Office Products (Office Prod), Grocery and Gourmet Food (Foods) and Video Games. From these four large data sets, eight data sets (four summary and four review data sets) were created for this study. The summary data sets were created by randomly selecting 1000 positive and 1000 negative summaries from a problem domain. Similarly, the review data sets were created by randomly selecting 1000 positive and 1000 negative reviews from a problem domain. For example, the created Cellphone review data set had reviews selected from the Amazon Cellphone data set only. Similarly, the created Foods summary data set had summaries selected from the Amazon Foods data set only.

These data sets were created since they represent different problem domains and contain a similar number of instances as compared to that presented in \cite{Pang:2002:ThumbsUp, PangLee:2004:ASentimentalEdu} and allow for a large number of experiments to be performed. The eight data sets used in this study are listed in table \ref{table:datasets}. Stop words and other irrelevant words were not removed from the data. For these to have no contribution to the overall sentiment, GASA must classify them as either `sentiment' or `amplifier' with a value of 0.

\begin{table}[H]
\centering
\caption{Data sets used in this study. Four review and four summary data sets were created. Each data set had 2000 instances.}
\label{table:datasets}
\begin{tabular}{ll}
\hline
\textbf{Data set} & \textbf{\begin{tabular}[c]{@{}l@{}}Number of positive/\\ negative instances\end{tabular}} \\ \hline
Cellphone         & 1000/1000 reviews                                                                         \\
Office Prod   & 1000/1000 reviews                                                                         \\ 
Foods             & 1000/1000 reviews                                                                         \\ 
Video Games       & 1000/1000 reviews                                                                         \\ 
Cellphone         & 1000/1000 summaries                                                                       \\ 
Office Prod   & 1000/1000 summaries                                                                       \\ 
Foods             & 1000/1000 summaries                                                                       \\ 
Video Games       & 1000/1000 summaries                                                                       \\ \hline
\end{tabular}
\end{table}

%----------------
% EXPERIMENTAL PARAMETERS
%---------------
\subsection{Experimental parameters}

The parameters used for the \textit{GASA} experiments are presented in table \ref{table:gasa-parameters}. These parameters were obtained by preliminary runs. The following section presents the results obtained by \textit{GASA} on several experiments.

\begin{table}[H]
\centering
\caption{\textit{GASA} parameters. These were obtained from preliminary runs.}
\label{table:gasa-parameters}
\begin{tabular}{ll}
\hline
\textbf{GASA Parameter}       & \textbf{Value} \\ \hline
Population Size               & 200            \\ 
Parent selection method       & Tournament     \\ 
Tournament size               & 7              \\ 
Maximum number of generations & 500            \\ 
Crossover rate                & 60\%             \\ 
Mutation rate                 & 40\%             \\ \hline
\end{tabular}
\end{table}

%------------------------------------------
% RESULTS
%------------------------------------------
\section{Results and Discussion}\label{results}

This section is split into three subsections. The purpose of the algorithm discussed the first subsection was to determine whether a given word was a sentiment or an amplifier word. The second subsection describes experiments whereby the goal was to determine the value of the sentiment word. And finally, the third subsection presents the results when \textit{GASA} was compared to other sentiment analysis algorithms. The results from the first two experiments demonstrate \textit{GASA}'s ability to generate a sentiment lexicon whereas the third experiment illustrates how \textit{GASA} performs as a sentiment analysis algorithm.

The amplifier dictionary was seeded with two words: ``not'' and ``never''; each had a value of ``-1'', representing negation. The sentiment dictionary was obtained from \cite{Hu:2004:MiningAndSummarizing} which we refer to as ``HuLiu-6786". This dictionary contains 6786 known sentiment words labelled as either positive or negative.

%------------------------------------------
% RESULTS: SENTIMENT OR AMPLIFIER
%------------------------------------------
\subsection{Predicting `sentiment' or `amplifier'}\label{sec:predict-sent-or-amp}

Several experiments were conducted whereby the classification problem was converted into a 2-class problem, namely sentiment and amplifier classes. Each word in the HuLiu-6786 dictionary was considered as a sentiment word.  Ten-fold cross-validation was used on the HuLiu-6786 dictionary whereby during each experiment, 9 folds from the dictionary were used for training, and the remaining fold from the dictionary was used for testing. \textit{GASA} had to predict whether each word in the test fold was a sentiment or an amplifier. The training and testing data was made up of dictionary words and not reviews or summary data.

\textit{GASA}'s fitness function did not take into consideration (during the evolutionary process) whether or not a training dictionary word was correctly classified as a sentiment or not. Thus, during the evolutionary process for these experiments, \textit{GASA} did not directly optimize the chromosomes in order to correctly distinguish between the two classes. Instead, \textit{GASA}'s goal was to correctly classify the overall sentiment of as many instances (reviews or summary data) as possible.

For these experiments, two data sets were created from the Cellphone, Office Products, Foods and Video Games data sets. Namely, all of the summary data combined and all of the review data combined. Thus, the two combined data sets had 8000 instances each. 

The results for these experiments are presented in tables \ref{table:2class-SentAmp-review} and \ref{table:2class-SentAmp-summary}. Each row in the table represents a particular word frequency condition; this is followed by the corresponding average number of dictionary words in the training data which met the word frequency condition and the average test accuracy across the 10-folds. Note that this test accuracy is in terms of the 2-class problem of distinguishing between a sentiment or amplifier word as described above. 

The word frequency condition is read as follows: a value of `$>10$' means that the experiment only took into consideration the training dictionary words which occurred at least 10 times in the review/summary data. In table 3, there were an average of 446 training dictionary words which had a frequency of 10. Similarly, a value of `$>250$' means that the experiment only took into consideration the training dictionary words which occurred at least 250 times in the review/summary data. There were an average of 23 words in the training data which had a frequency of 250. The condition `$>0$' implied that a dictionary word had to occur at least once in the review/summary data. 

The purpose of using the word frequency condition was to determine the effect on the number of times a word was present in the training data and \textit{GASA}'s ability to correctly classify the words in the test set which also had such a frequency.

\begin{table}[H]
\caption{Test accuracy (\%) results on the two class problem (sentiment and
amplifier) on all of the review data combined into a single data set.
Ten-fold cross-validation was used. \label{table:SvA-2}} \label{table:2class-SentAmp-review}
\begin{tabular}{>{\centering}p{2cm}>{\centering}p{3cm}>{\centering}p{2cm}}
\hline 
\textbf{Word Frequency} & \textbf{Total number of dictionary words} & \textbf{Accuracy (\%)}\tabularnewline
\hline 
$>0$ & 1917 & 51.39\tabularnewline
$>10$ & 446 & 54.80\tabularnewline
$>15$ & 334 & 55.47\tabularnewline
$>20$ & 260 & 56.66\tabularnewline
$>100$ & 56 & 64.52\tabularnewline
$>250$ & 23 & 72.00\tabularnewline
\hline 
\end{tabular}
\end{table}

\begin{table}[H]
\caption{Test accuracy (\%) results on the two class problem (sentiment and amplifier)
on all of the summary data combined into a single data set. Ten-fold
cross-validation was used. \label{table:SvA-1}} \label{table:2class-SentAmp-summary}
\begin{tabular}{>{\centering}p{2cm}>{\centering}p{3cm}>{\centering}p{2cm}}
\hline 
\textbf{Word Frequency} & \textbf{Total number of dictionary words} & \textbf{Accuracy (\%)}\tabularnewline
\hline 
$>0$ & 626 & 55.11\tabularnewline
$>10$ & 76 & 68.42\tabularnewline
$>15$ & 56 & 67.86\tabularnewline
$>20$ & 41 & 75.61\tabularnewline
\hline 
\end{tabular}
\end{table}

When all of the words are taken into consideration, i.e. a frequency value `$>0$' , \textit{GASA} achieved an accuracy of 51.39\% and 55.11\% on the review and summary data respectively.  The accuracy improved when the word frequency condition was increased. In terms of the review data, the accuracy went from 51.39\% to 72.00\% when the word frequency was increased from `greater than 0' to `greater than 250'. 

For the combined summary data, when the words had a frequency of at least 20 the accuracy was 75.61\% as opposed to an accuracy of 55.11\% for a frequency condition greater than 0. The combined review data set had more words than the combined summary data set because the summaries are short text. For this reason, the conditions were stopped at 20 for the combined summary data. Words from the dictionary which occur with a small frequency are more challenging for \textit{GASA} to correctly classify  as a sentiment or amplifier since they occur infrequently in the data. Nonetheless, the findings reveal that \textit{GASA} is able to extend a sentiment and amplifier lexicon provided that the words occur with a large frequency in the training data.

%------------------------------------------
% RESULTS: VALUE OF SENTIMENT
%------------------------------------------
\subsection{Predicting the value of the sentiment}

A set of experiments was conducted in order to determine how effective \textit{GASA} would be at classifying the sentiment value of a set of words instead of sentences. In order to achieve this, the HuLiu-6786 dictionary was used, and a certain percentage of the words in the dictionary were considered as unknown. The problem was converted into a 2-class classification problem, namely positive and negative sentiment values. Thus, in terms of the classification-value pair, only the ``value'' aspect was taken into consideration.

The HuLiu-6786 dictionary contained more sentiment words than were present in the data sets, and thus only the dictionary words found in the training data sets were considered --- this set was named \textit{S}.  

Ten-fold cross-validation was used in the following manner: 10 folds were randomly created from \textit{S}, 9 folds were seeded into \textit{GASA} and the algorithm was executed as defined in section \ref{Proposed}. At the end of the GA generational loop the algorithm had to predict the sentiment value of the words in the test fold as either ``positive'' or ``negative''.  The predictions were then compared against the correct values in order to determine the accuracy. Similarly to the experiment described in subsection \ref{sec:predict-sent-or-amp}, \textit{GASA} did not directly optimize the chromosomes in order to correctly distinguish between the two classes. \textit{GASA}'s objective was to correctly classify the overall sentiment of as many instances (reviews or summary data) as possible.

\begin{table}[H]
\caption{Test accuracy (\%) results on the two class problem (positive and negative sentiment)
on all of the review data combined into a single data set. Ten-fold
cross-validation was used.} \label{table:2class-PosNeg-review}
\begin{tabular}{>{\centering}p{2cm}>{\centering}p{3cm}>{\centering}p{2cm}}
\hline 
\textbf{Word Frequency} & \textbf{Total number of dictionary words} & \textbf{Accuracy (\%)}\tabularnewline
\hline 
$>0$ & 1917 & 36.62\tabularnewline
$>10$ & 446 & 44.39\tabularnewline
$>15$ & 334 & 46.71\tabularnewline
$>20$ & 260 & 49.23\tabularnewline
$>100$ & 56 & 69.64\tabularnewline
$>250$ & 23 & 82.61\tabularnewline
\hline 
\end{tabular}
\end{table}

\begin{table}[H]
\caption{Test accuracy (\%) results on the two class problem (positive and negative sentiment)
on all of the summary data combined into a single data set. Ten-fold
cross-validation was used.} \label{table:2class-PosNeg-summary}
\begin{tabular}{>{\centering}p{2cm}>{\centering}p{3cm}>{\centering}p{2cm}}
\hline 
\textbf{Word Frequency} & \textbf{Total number of dictionary words} & \textbf{Accuracy (\%)}\tabularnewline
\hline 
$>0$ & 626 & 53.04\tabularnewline
$>10$ & 76 & 86.84\tabularnewline
$>15$ & 56 & 92.86\tabularnewline
$>20$ & 41 & 95.12\tabularnewline
\hline 
\end{tabular}
\end{table}

For these experiments, the same data sets which were described in subsection \ref{sec:predict-sent-or-amp} were used.  The results for these experiments are presented in tables \ref{table:2class-PosNeg-review} and \ref{table:2class-PosNeg-summary}. Subsection \ref{sec:predict-sent-or-amp} describes how to interpret the tables. Dictionary words in the combined summary data set which had a frequency value of at least 20 resulted in an accuracy of 95.12\%. Sentiment words had to appear a greater number of times in the combined review data set in order to achieve a higher accuracy; more precisely, words which had a frequency of at least 250 times resulted in an accuracy of 82.61\%. These results reveal that \textit{GASA} is able to extend a sentiment lexicon provided that the words occur frequently.

%------------------------------------------
% RESULTS: GASA VS OTHER
%------------------------------------------
\subsection{Comparison of GASA with commercial Sentiment Tools}

How good is \textit{GASA}? To check we compared \textit{GASA} seeded with the HuLiu-6786 dictionary to other sentiment analysis methods including AlchemyAPI \cite{alchemyapi}, MeaningCloud \cite{meaningcloud}, NLTK \cite{nltk}, Lexalytics \cite{lexalytics}, LingPipe \cite{lingpipe}, Stanford sentiment analysis \cite{Socher:2013:Recursive, Manning:TheStanford:2014}, SentiStrength \cite{Thelwall:2010:Sentiment, Thelwall:2012:Sentiment} and Dandelion API \cite{dandelion}. AlchemyAPI, MeaningCloud, Lexalytics and Dandelion are commercial APIs. LingPipe and SentiStrength have both commercial and non-commercial licences. The results of the comparison are presented in table \ref{table:22}. 

In terms of the summary data, the top 3 ranking methods in order of performance were LingPipe, AlchemyAPI and \textit{GASA} with an average test accuracy of 77.75\%, 72.88\% and 69.92\% respectively.  When comparing \textit{GASA} to the four commercial APIs, AlchemyAPI achieved the best accuracy, while \textit{GASA} outperformed Dandelion, Lexalytics and MeaningCloud. 

In terms of the review data sets, the top three performing methods were LingPipe, AlchemyAPI and SentiStrength. \textit{GASA} was outperformed by two commercial API, namely AlchemyAPI and Dandelion. \textit{GASA} achieved higher test accuracy when compared to two commercial APIs, namely, Lexalytics and MeaningCloud. 

Appendices A and B illustrates several examples of the review data used. The predicted sentiment on the sample reviews from \textit{GASA} and other sentiment analysis methods is presented. The reviews were randomly selected in order to illustrate cases where \textit{GASA} correctly and incorrectly classified the sentiment.

%--------------------------------------------------------
% GASA-HuLiu vs Other Methods
%--------------------------------------------------------

\begin{table*}[t]

\caption{Test accuracy (\%) illustrating a comparison between other commercial and non commercial sentiment analysis methods and GASA. The 70/30 holdout method was used, and all of
the data sets had a size of 2000. Types ``S'' and ``R'' denote summary and review data sets respectively. Watson refers to Alchemy API, MC to MeaningCloud, LEX to Lexalytics, LP to LingPipe, SS to SentiStrength, DL to Dandelion and SD to Stanford sentiment analysis. \label{table:22}}

\begin{centering}
\begin{tabular}{l>{\centering}p{7mm}>{\centering}p{11mm}>{\centering}p{8mm}>{\centering}p{9mm}>{\centering}p{8mm}>{\centering}p{8mm}>{\centering}p{8mm}>{\centering}p{8mm}>{\centering}p{8mm}>{\centering}p{2cm}}
\hline 
\textbf{Data set} & \textbf{Type} & \textbf{Watson} & \textbf{MC} & \textbf{NLTK} & \textbf{LEX} & \textbf{LP} & \textbf{SS} & \textbf{DL} & \textbf{SD} & \textbf{GASA}\tabularnewline
\hline 
Cellphone & S & 75.67 & 60.00 & 60.67 & 37.50 & 80.17 & 68.00 & 55.17 & 54.83 & 74.33\tabularnewline
Office Prod & S & 71.50 & 59.83 & 59.83 & 39.83 & 80.00 & 69.83 & 53.50 & 55.83 & 70.67\tabularnewline
Foods & S & 68.83 & 52.67 & 57.67 & 40.00 & 74.50 & 64.50 & 50.17 & 51.00 & 63.33\tabularnewline
Video Games & S & 75.50 & 59.83 & 57.67 & 38.33 & 76.33 & 66.17 & 55.00 & 51.33 & 71.33\tabularnewline
\hline 
\multicolumn{2}{l}{\textbf{Average Summary}} & \emph{72.88 } & \emph{58.08 } & \emph{58.96 } & \emph{38.92 } & \emph{77.75 } & \emph{67.13} & \emph{53.46} & \emph{53.25} & \emph{69.92 }\tabularnewline
\hline 
Cellphone & R & 70.17 & 62.33 & 64.17 & 58.50 & 81.50 & 67.67 & 65.00 & 54.17 & 69.17\tabularnewline
Office Prod & R & 71.50 & 64.83 & 64.83 & 60.17 & 81.50 & 72.00 & 68.33 & 53.67 & 68.67\tabularnewline
Foods & R & 71.17 & 62.83 & 63.83 & 64.00 & 78.83 & 70.67 & 69.67 & 52.83 & 64.83\tabularnewline
Video Games & R & 69.17 & 62.00 & 71.83 & 53.00 & 79.17 & 68.17 & 67.00 & 58.17 & 66.00\tabularnewline
\multicolumn{2}{l}{\textbf{Average Review}} & \emph{70.50} & \emph{63.00} & \emph{66.17} & \emph{58.92} & \emph{80.25} & \emph{69.63} & \emph{67.50} & \emph{54.71} & \emph{67.17}\tabularnewline
\hline 
\end{tabular}
\par\end{centering}

\protect
\end{table*}

\section{Extending GASA (CA-GASA)}

When determining the classification for an unknown word \textit{w}, the \textit{GASA} algorithm does not take into consideration the words before and after \textit{w}, i.e. it is context independent. This ignores the fact that many words have different meanings --- with different sentiments. How can we begin to allow for multiple, context-dependent sentiments? We propose to allocate a context-dependent classification to an unknown word \textit{w}; an approach we call Context Aware GASA \textit{(CA-GASA)}. In order to achieve this, several modifications to \textit{GASA} are required. The primary modification lies within the representation of the chromosomes.

\begin{figure}[!ht]
  \centering
          \includegraphics[width=0.48\textwidth]{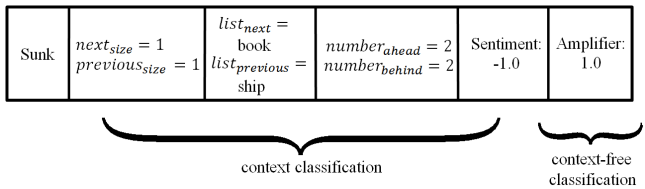}
  \caption{Example of a \textit{CA-GASA} chromosome.}
\label{fig:cagasachromo}
\end{figure}

Each gene contains two principle parts, the context classification and the context-free classification. When a word in an instance of data is evaluated the classification-value pair is obtained from either the context classification or context-free classification. Two lists of words are used in order to make this decision, namely $list_{next}$ and $list_{previous}$. When a \textit{CA-GASA} chromosome is evaluated on an instance of data \textit{i} on a word \textit{w}, the context classification-value pair is allocated if the word \textit{w} is surrounded by the words in $list_{next}$ and $list_{previous}$. If this is not the case, then the context-free classification-value pair is allocated. This process is further discussed below. Figure \ref{fig:cagasachromo} illustrates an example of a \textit{CA-GASA} chromosome. 

In order to enable multiple classification-value pairs to be associated with a word, a new gene encoding is used. For a word \textit{w}, each gene has the following properties:

\begin{itemize}
  \item The word \textit{w}.
  \item The context rule which is defined as follows:
  \begin{itemize}
	\item The maximum possible size of the next context words, denoted as $next_{size}$.
	\item The maximum possible size of the previous context words, denoted as $previous_{size}$.
	\item The list of the next context words, denoted as $list_{next}$.
	\item The list of the previous context words, denoted as $list_{previous}$.
	\item The number of words to look ahead and compare with $list_{next}$ , denoted as $number_{ahead}$.
	\item The number of words to look behind and compare with $list_{previous}$, denoted as $number_{behind}$.
	\item The context classification-value pair.
  \end{itemize}
  \item The context-free classification-value pair.
\end{itemize}

\subsection{CA-GASA chromosome evaluation}\label{sub:chromoEval}

Assume that a \textit{CA-GASA} chromosome \textit{c} is being applied to an instance of data. A word \textit{w} in the sentence is evaluated as follows.  Starting from the word \textit{w}, look at the next $number_{ahead}$ words from \textit{w} and add them to the $list_x$. Once again, starting from the word \textit{w}, look at the previous $number_{behind}$ words from \textit{w} and add them to the $list_y$.

Let the size of $list_x$ be denoted as $size_x$, and let the size of $list_y$ be denoted as $size_y$. Let the number of words in the intersection between $list_x$ and $list_{next}$ be denoted by \textit{a}, and let the number of words in the intersection between $list_y$ and $list_{previous}$ be denoted by \textit{b}.

If $\frac{a+b}{size_x + size_y} \ge 0.5$, then the word \textit{w} is classified by the context classification value. Otherwise, the word \textit{w} is classified by the context-free classification value.

The \textit{CA-GASA} chromosome in figure \ref{fig:cagasachromo}  has one next context word, ``ship'', and one previous context word, ``book''. Assume the sentence ``the ship sunk'' is being evaluated, and the classification-value for the word ``sunk'' is being determined. In this case $list_x$ is empty, and   $list_y$ = \{ship, the\}. Consequently, $size_x = 0$ and $size_y = 2$. The intersection between $list_x$ and $list_{next}$  is empty, and the intersection between  $list_y$ and $list_{previous}$ is \{ship\}, and thus, \textit{a} = 0 and \textit{b} = 1. The word ``sunk'' is classified by the context classification-value since $\frac{0+1}{0+2} \ge 0.5$, i.e. ``sunk'' is classified as a sentiment word with a value of -1.0 (since the context classification-value pair in the figure is a sentiment with value of -1.0).

\subsection{CA-GASA Results}

From table \ref{table:cagasa-gasa-comparison-results}, when HuLiu-6786 was seeded into the proposed methods, it is observed that \textit{GASA} outperformed \textit{CA-GASA} on 2 summary data sets, and tied in the other data sets, whereas \textit{CA-GASA} outperformed \textit{GASA} on 3 of the review data sets. One drawback of \textit{CA-GASA} is that the search space is significantly larger than \textit{GASA} and as a result the training time is much longer. Given this drawback, \textit{CA-GASA} was not tested against the other sentiment analysis algorithms.

In order to address the large training time, it would be of interest to determine if an approach could be proposed in order to find which words in some data set have more than one meaning, and to create the context classification for those words only. This would reduce the complexity of \textit{CA-GASA} whilst retaining the ability to perform word disambiguation. Unknown words which only express one sentiment regardless of the context could be represented using \textit{GASA}, and words which have more than one sentiment could use the \textit{CA-GASA} representation.

\begin{table}
\caption{Test accuracy (\%) on the summary and review data showing a comparison
between GASA and CA-GASA. Ten-fold cross-validation was used, and
all of the data sets had a size of 2000. Both GASA and CA-GASA were
seeded with the HuLiu-6786 sentiment dictionary.} \label{table:cagasa-gasa-comparison-results}
\begin{tabular}{l>{\centering}p{10mm}>{\centering}p{10mm}>{\centering}p{10mm}>{\centering}p{10mm}}
\hline 
 & \multicolumn{2}{>{\centering}p{22mm}}{\textbf{Summary Data}} & \multicolumn{2}{>{\centering}p{22mm}}{\textbf{Review Data}}\tabularnewline
\hline 
\textbf{Data set} & \textbf{GASA} & \textbf{CA-GASA} & \textbf{GASA} & \textbf{CA-GASA}\tabularnewline
\hline 
Cellphone & 73.55 & 71.75 & 69.10 & 70.05\tabularnewline
Office Prod & 71.75 & 70.70 & 71.00 & 68.20\tabularnewline
Foods & 65.75 & 65.75 & 67.15 & 67.35\tabularnewline
Video Games & 69.40 & 69.40 & 65.45 & 67.00\tabularnewline
\hline 
\end{tabular}
\end{table}

%--------------------------------------------------------
% Conclusion
%--------------------------------------------------------
\section{Conclusion}\label{conclusion}

Being able to determine the sentiment of text is a useful ability to businesses and other entities in order to gain an understanding of people's opinions on their products and services. This study proposed a GA approach in order to classify the sentiment of sentences by optimizing unknown words as either a sentiment or an amplifier word. This study proposes a way to represent the sentiments and amplifiers through the use of simple mathematical expressions in order to evaluate the final sentiment of sentences. The experimental results revealed that \textit{GASA} was able to outperform certain commercial APIs.

One advantage of \textit{GASA} is that the algorithm can grow a sentiment dictionary which can be reused or further improved upon. The experiments suggested that if a particular word appeared a large number of times in the training data set, then the proposed method is likely to correctly classify its sentiment. 

We also proposed \textit{CA-GASA}, and the rationale behind this modification was to provide the ability to allocate a sentiment based on the context for which the words are in. This method requires additional work in order to reduce the complex search space. It would also be of interest in order to investigate if an ensemble of \textit{GASA} chromosomes could outperform the accuracy of a single one. 

%--------------------------------------------------------
% Appendix
%--------------------------------------------------------
\appendix

%------------------------------------
% CORRECTLY CLASSIFIED BY GASA
%------------------------------------

\section{Correctly Classified by \textit{GASA}}\label{appendixA}

This section presents a random sample of the reviews which were correctly classified by \textit{GASA}. The reviews were obtained from the Video Game data set. The reviews were lemmatized using Stanford CoreNLP \cite{Manning:TheStanford:2014}. The sentence ``we thought this movie was quite entertaining'' is lemmatized as follows: ``we think this movie be quite entertaining''. The results are compared to the following four sentiment analysis entities: NLTK, Alchemy API, SentiStrength and MeaningCloud. The reviews are as follows:

\textbf{Text:} ``\textit{have purchase the original when release several year ago, and thoroughly enjoy it, I be excite to see a new version out for 2006. it live up to the challenge I expect from this game. it be difficult to find truly challenging puzzle game, and you will not be disappoint with this. my only disappoint come after solve it as the original provide a video of the programmer, although this version do offer the opportunity to replay and select from several final outcome.}''

\textbf{Result:} Correctly classified by \textit{GASA} as positive.

\textbf{Text:} ``\textit{have not finish it yet, but it sure be a lot of fun! yes, there be/will be even better game. and yes, maybe GTA V be better (it have at least three time more budget so no surprise there). but Watch Dogs be truly a great game! some have unrealistically high expectation or be real hater and should just stick to buy gta every five year instead of buy each game and then give they bad review over and over... before write another obvious review: yes, WE all KNOW that ``gta5'' be NOT ``watch dog'' ``call of duty'' be NOT ``candy crush'' and ``Far Cry 3'' be NOT ``Tetris''''}

\textbf{Result:} Correctly classified by \textit{GASA} as positive.

\textbf{Text:}  ``\textit{I recommend this game. very little communication loss or problem. if problem occur they be very good as about replace what you lose. the game be fun to play.}''

\textbf{Result:} Correctly classified by \textit{GASA} as positive. 

\textbf{Text:} ``\textit{I buy this game because I remember how Madden use to be. I have hear the gameplay be break (it be, horribly), but I think, hey, if I get use to the stupid game glitch at least it will look good. no. wow be I wrong. everything about this game be atrocious . the only thing that look crisp in this game be the score, which incidentally be the only reason I know it be display in hd. every time a player score a touchdown he literally run through the stand and disappear into the mesh. I could go on, but just think that I pay more than \$ 5 for this game be make I extremely angry right no.}''

\textbf{Result:} Correctly classified by \textit{GASA} as negative.

\textbf{Text:} ``\textit{I play game on a desktop pc in my office, and on a laptop in the living room. I also frequently rebuild my machine and upgrade hardware. I be interested in the online community that will follow this game, however I will not support such strict measure in copy protection. I have cancel my pre-order and may purchase later once this limited activation process be remove.}''

\textbf{Result:} Correctly classified by \textit{GASA} as negative.

%------------------------------------
% INCORRECTLY CLASSIFIED BY GASA
%------------------------------------

\section{Incorrectly Classified by \textit{GASA}}\label{appendixB}

This section presents a random sample of the reviews which were incorrectly classified by \textit{GASA}. Refer to appendix A for details regarding the data, lemmatization and the algorithms used for comparison. The reviews are as follows:

\textbf{Text:} ``\textit{I like final Fantasy 13 but it just do not draw I in that well and the story be slightly confusing. with that be say I absolutely love final Fantasy 13-2! the story be weird because of time travel, but I think it be awesome. they seem to have fix everything about the first one. I also like the fact that this game have multiple ending and new game plus! buy it!}''

\textbf{Result:} The correct classification for the review is positive. \textit{GASA} classified it as negative. Three of the algorithms correctly classified the text as positive, and one algorithm classified it as negative.

\textbf{Text:} ``\textit{outstanding shooter game. it be set in world war two and you better not run out in the open. good graphic and story.}''

\textbf{Result:} The correct classification for the review is positive. \textit{GASA} classified it as negative. Only one algorithm correctly classified the text. Two algorithms classified it as neutral and the remaining algorithm classified it as negative.

\textbf{Text:}``\textit{they sit a little bit loose on the controller's peg, so there be some extra play. not shabby, not stellar. put some more life into a controller that you think you be go to lose.}''

\textbf{Result:} The correct classification for the review is positive. \textit{GASA} classified it as negative. All of the algorithms incorrectly classified the text as negataive. 

\textbf{Text:} ``\textit{this headset rock, that be all you need to know. unless you need more, here be some pro's and con's. pro's 1. affordable price2. long cable3. lightweight design4. adjustable microphone5. Chat Boost / Independent Game and Chat Sound6. stereo ExpanderCon's 1. slightly complicated set - up2. cheap ear Cushions The hiss sound people complain about be negligible, it disappear after just a few minute of use and be otherwise drown out by game. amazing quality and durability at a affordable price, highly recommend. UPDATE : 2/18/11 the microphone serve I very well for only 17 day before the Right Earphone cease to function. a day later, the right earphone completely fall off. I get off recommend the product initially, but the fact that the headset can stop work so quickly be terrible. I be return these for a refund and be go to try another headset. I no longer recommend these headphone.}''

\textbf{Result:} The correct classification for the review is negative. \textit{GASA} classified it as positive. Three of the algorithms correctly classified the text as negative whereas the other classified it as neutral.

\textbf{Text:}``\textit{Brunswick pro Bowling be not worth much. game be slow, unnatural, and poor scripting. I can not believe someone can not make a better program then the Wii bowling which be do pretty good. do not waste you money. I want sport game that actually feel and act like real sport. how hard can that be?}''

\textbf{Result:} The correct classification for the review is negative. \textit{GASA} classified it as positive. Three of the algorithms correctly classified the text as negative whereas the other classified it as positive.

\begin{acks}
The financial assistance of the National Research Foundation (NRF) towards this research is hereby acknowledged. Opinions expressed and conclusions arrived at, are those of the author and are not necessarily to be attributed to the NRF. The computations were performed at the University of Geneva on the Baobab cluster. The authors would like to thank Alchemy API, Dandelion API, Lexalytics, MeaningCloud and SentiStrength for granting access to their APIs. The authors would like to thank Etienne Vos for his feedback and comments.
\end{acks}

\bibliographystyle{ACM-Reference-Format}

\end{document}